\documentclass[letterpaper, 10 pt, conference]{ieeeconf}  % Use this line for US letter paper

\IEEEoverridecommandlockouts
\usepackage{amsmath}
\usepackage{amssymb}
\usepackage{graphicx}
\usepackage{float}
\usepackage{booktabs}
\usepackage{multirow}
\usepackage{tabularx}
\usepackage{placeins}
\usepackage{cite}
\usepackage{caption}
\usepackage{hyperref}
\usepackage{tikz}
\usepackage{scalerel}
\usepackage{tikz}
\usetikzlibrary{svg.path}
\definecolor{orcidlogocol}{HTML}{A6CE39}
\tikzset{
    orcidlogo/.pic={
        \fill[orcidlogocol] svg{M256,128c0,70.7-57.3,128-128,128C57.3,256,0,198.7,0,128C0,57.3,57.3,0,128,0C198.7,0,256,57.3,256,128z};
        \fill[white] svg{M86.3,186.2H70.9V79.1h15.4v48.4V186.2z}
        svg{M108.9,79.1h41.6c39.6,0,57,28.3,57,53.6c0,27.5-21.5,53.6-56.8,53.6h-41.8V79.1z M124.3,172.4h24.5c34.9,0,42.9-26.5,42.9-39.7c0-21.5-13.7-39.7-43.7-39.7h-23.7V172.4z}
        svg{M88.7,56.8c0,5.5-4.5,10.1-10.1,10.1c-5.6,0-10.1-4.6-10.1-10.1c0-5.6,4.5-10.1,10.1-10.1C84.2,46.7,88.7,51.3,88.7,56.8z};
    }
}
\newcommand\orcidicon[1]{\href{https://orcid.org/#1}{\mbox{\scalerel*{
    \begin{tikzpicture}[yscale=-1,transform shape]
    \pic{orcidlogo};
    \end{tikzpicture}
}{|}}}}

\title{\LARGE \bf
DAMM-LOAM: Degeneracy Aware Multi-Metric LiDAR Odometry and Mapping
}
\author{Nishant Chandna\,\orcidicon{0009-0008-0423-3641}$^{1,*}$, 
        Akshat Kaushal\,\orcidicon{0009-0001-4409-0429}$^{1,*}$%
\thanks{*These authors contributed equally to this work.}%
    \thanks{$^{1}$\mbox{Unmanned Aerial Systems-Delhi Technological University},
    \texttt{\{nishantchandna\_23ch049, akshatkaushal\_23me036\}@dtu.ac.in}}%
}

\begin{document}

\maketitle
\thispagestyle{empty}
\pagestyle{empty}

% Begin content here...

%%%%%%%%%%%%%%%%%%%%%%%%%%%%%%%%%%%%%%%%%%%%%%%%%%%%%%%%%%%%%%%%%%%%%%%%%%%%%%%%
\begin{abstract}

LiDAR Simultaneous Localization and Mapping (SLAM) systems are essential for enabling precise navigation and environmental reconstruction across various applications. Although current point-to-plane ICP algorithms perform effectively in structured, feature-rich environments, they struggle in scenarios with sparse features, repetitive geometric structures, and high-frequency motion. This leads to degeneracy in 6-DOF pose estimation. Most state-of-the-art algorithms address these challenges by incorporating additional sensing modalities, but LiDAR-only solutions continue to face limitations under such conditions. To address these issues, we propose a novel Degeneracy-Aware Multi-Metric LiDAR Odometry and Mapping (DAMM-LOAM) module. Our system improves mapping accuracy through point cloud classification based on surface normals and neighborhood analysis. Points are classified into ground, walls, roof, edges, and non-planar points, enabling accurate correspondences. A Degeneracy-based weighted least squares-based ICP algorithm is then applied for accurate odometry estimation. Additionally, a Scan Context based back-end is implemented to support robust loop closures. DAMM-LOAM demonstrates significant improvements in odometry accuracy, especially in indoor environments such as long corridors.
\end{abstract}

%%%%%%%%%%%%%%%%%%%%%%%%%%%%%%%%%%%%%%%%%%%%%%%%%%%%%%%%%%%%%%%%%%%%%%%%%%%%%%%%
\section{INTRODUCTION}

LiDAR Odometry and Mapping has achieved remarkable accuracy in recent years. With a wide range of applications in various fields, including robot navigation, construction\cite{construction}, autonomous driving\cite{autonomous_navigation}, and underground mining exploration\cite{mining_underground}, it continues to be a fundamental technology for robust SLAM in complex and dynamic environments. LiDAR sensors are capable of providing stable and rich geometric features for spatial understanding, enabling high versatility in various environments, even under changing lighting conditions. The Iterative Closest Point (ICP) algorithm is the most commonly used method for scan matching and pose estimation. 

\begin{figure}
    \centering
    \includegraphics[width=1\linewidth]{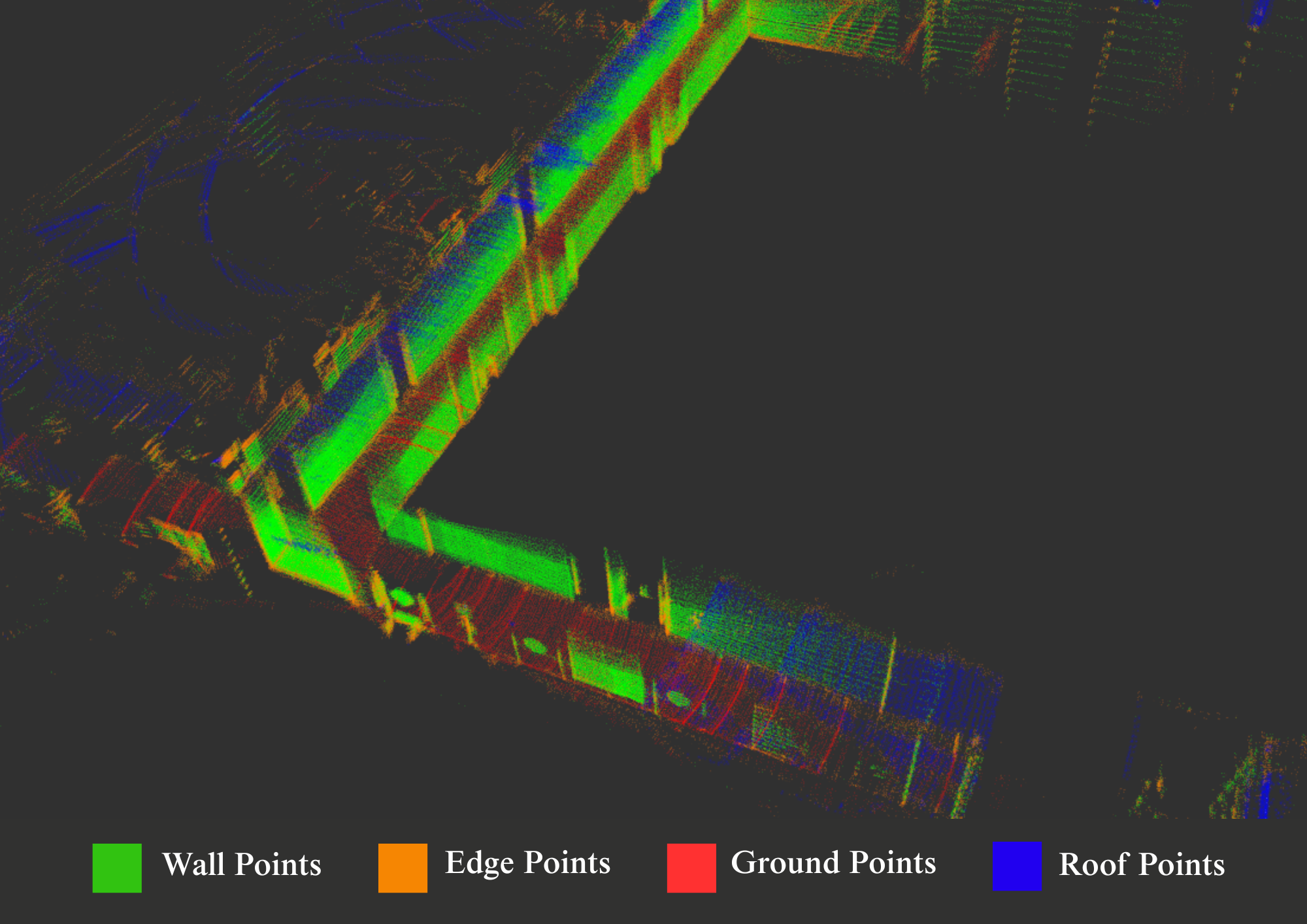}
    \caption{The image demonstrates feature classification and mapping accuracy of DAMM-LOAM on Ground Challenge Corridor 1 Dataset}
    \label{fig:enter-label}
\end{figure}
In recent years, multiple algorithms have used point-to-point ICP or point-to-plane ICP \cite{KISS-ICP}\cite{plane}\cite{cticp}\cite{xicp}. These algorithms show great performance in specific environments with specific parameters, but fail to perform well in varying dominant geometries. Some perform exceptionally well in outdoor environments but fail in indoor environments, highlighting a common domain-specific performance limitation.

However, the accuracy of LiDAR-based solutions is highly dependent on the presence of sufficient geometric features in the environment. Different geometric features have different impacts on the ICP solution. In scenarios where such constraints are limited, such as in hallways, underground tunnels, or long corridors, the underlying optimization problem often becomes ill-conditioned, resulting in degraded or inaccurate localization performance. To address these challenges, several approaches have been proposed to enhance degeneracy robustness \cite{mad}\cite{xicp}\cite{robustrank}. Some methods focus on improving point correspondences, while others actively detect degenerate scenarios and adapt the estimation process either by constraining motion along poorly observable directions or fusing complementary sensor modalities. Recent works have explored strategies such as leveraging multi-modal fusion to compensate for environments with limited perceptual features, and adaptively adjusting solver behaviour based on observability analysis. 
\\
In this work we introduce two main contributions:

\begin{enumerate}
    \item \textbf{Normal Map based Semantic Feature Extraction:} \\
    We introduce a method to classify surface features, namely \textit{ground}, \textit{walls}, \textit{roof}, and \textit{edges} using a normal map. These semantic labels are then leveraged to perform multi-metric correspondence search and improve the robustness of ICP registration.

    \item \textbf{Degeneracy-Aware Point Wise Adaptive Weighting:} \\
    We develop a point-wise adaptive weighting scheme based on degeneracy analysis. These weights are integrated into the weighted least squares objective of ICP, allowing optimization to account for spatial observability and accuracy in under-constrained environments.
\end{enumerate}

\section{RELATED WORK}

\subsection{LiDAR Odometry}

LiDAR odometry forms the basis of many contemporary SLAM systems. Owing to its high precision, long-range sensing, and robustness to lighting variations, it has become a prominent focus of recent research \cite{KISS-ICP}\cite{genz}\cite{gicp}\cite{loam}. 
 
Some SLAM methods adopt varied point selection strategies: some apply uniform voxel grid down-sampling to reduce computational load \cite{loam}, while others extract semantic or geometric features to retain structurally informative points \cite{mulls}\cite{superodom}. Different types of ICP algorithms have been employed in various studies. KISS-ICP \cite{KISS-ICP} leverages only point-to-point ICP. While it shows exceptional performance compared to other algorithms, it fails in structured and degenerate environments such as long corridors due to its reliance on a single error metric. In contrast, other algorithms \cite{superodom} \cite{mulls}, on the other hand, combine combinations of point-to-point, point-to-plane and point-to-line error metrics to remain robust under such conditions. GenZ-ICP\cite{genz} takes it one step further by implementing adaptive weighting based on the number of planar and non-planar points, improving accuracy and performance in both indoors and outdoors. 

While the classification of planar and non-planar regions shows strong performance, it does not fully exploit all the available information. Integrating semantic information into LiDAR SLAM marks a significant advancement toward developing more robust and accurate LiDAR odometry systems \cite{mulls}. Our work focuses on classifying point cloud based on information obtained from a projected normal map, and hence applying multi-metric point cloud registration.

\subsection{Feature Extraction}
Over the years, multiple studies have proposed methods for feature extraction in structural element classification \cite{mulls}, \cite{feature}, \cite{pca_feature}. Techniques like Principal Component Analysis (PCA), Region Growing Segmentation \cite{region_growing}, RANSAC-based plane detection\cite{ransac}. PCA remains a pivotal technique for normal vector estimation and feature extraction in point cloud. SuperOdometry\cite{superodom} uses PCA to distinguish and extract planar, non-planar, and edge features from the point cloud data. PCA-based methods estimate surface normals by analysing eigenvectors and eigenvalues of covariance matrices. Accuracy of PCA decomposition heavily relies on the correct selection of neighbourhood, and hence it requires highly tuned values for different LiDARs. NV-LIOM \cite{NV-LIOM} uses a spherical projection-based normal extraction method. For every pixel in the range image, the normal vector is computed by calculating local derivatives using a window-based approach. Inspired by this, we further utilised this normal map to classify structural geometry into ground plane, walls, roof, edges and non-planar points. Spherical projection methods generally have lower computational load compared to both RANSAC and PCA. 
\subsection{Degeneracy Awareness}
Point cloud registration methods, such as ICP, are widely adopted and have shown strong performance across a variety of real-world applications. However, they are prone to degeneracy during optimization, especially in environments with poor geometric structure or repetitive patterns—such as tunnels or corridors, which can result in ill-conditioning along one or more degrees of freedom. 

Extensive research has been conducted on identifying degeneracy in such environments. Some approaches, like those in \cite{dare-slam}\cite{lion}, utilise the condition number defined as the ratio of the largest to the smallest eigenvalue of the optimization Hessian as a single metric to detect degeneracy across all six degrees of freedom.  In contrast, the method in \cite{dams} evaluates each eigenvalue individually to classify the corresponding direction as degenerate or not.

To address degeneracy, various solutions have been proposed. Some methods, such as \cite{fus1}\cite{fus2}, rely on external odometry sources when the environment becomes degenerate. Others, including \cite{xicp}\cite{cticp}, enhance the registration process in degenerate scenarios by reformulating the optimization with additional soft or hard constraints that limit movement along poorly observable directions.

\begin{figure*}[t]
    \centering
    \vspace{1mm}
    \includegraphics[width=\linewidth]{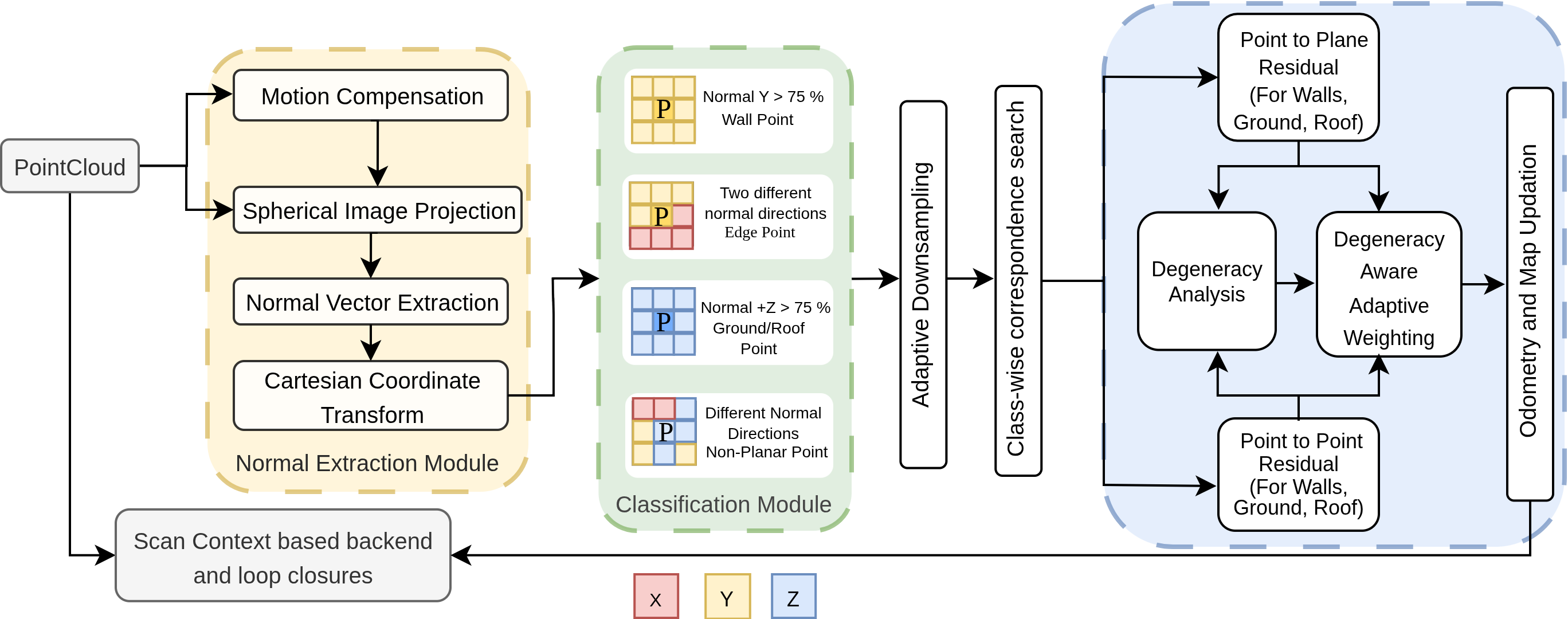}
    \caption{The pipeline extracts surface normals from LiDAR point cloud, classifies points into structural elements, and performs robust, degeneracy-aware odometry and mapping with loop closure using Scan Context for global consistency.}
    \label{fig:damm-loam}
\end{figure*}
\section{SYSTEM OVERVIEW}

Aligning a source point cloud ($\mathbf{S}$) to a reference point cloud $\mathbf{R}$ using a rigid body transformation $T\in\text{SE}(3)$ is a fundamental problem in LiDAR odometry, typically addressed by algorithms like Iterative Closest Point (ICP). While ICP aims to estimate the optimal transformation by minimizing residual error, its efficacy depends on robust data association and accurate error formulation. 

This section introduces DAMM-LOAM, a novel LiDAR odometry framework designed to enhance accuracy and robustness through a comprehensive, multi-stage pipeline. An overview of the process is shown in fig. \ref{fig:damm-loam}.

DAMM-LOAM processes LiDAR point cloud for odometry and mapping. It begins with point cloud classification, segmenting the source cloud into geometric categories: \emph{wall}, \emph{edge}, \emph{ground}, \emph{roof}, and \emph{non-planar}. This enables adaptive processing and tailored correspondence finding. Following classification, points are adaptively downsampled to reduce computation while preserving geometric information.

Class-wise correspondences are then established using a voxel hash map. Degeneracy analysis identifies and mitigates ill-posed configurations. Finally, joint optimization combines point-to-point and point-to-plane residuals to estimate the optimal pose transformation $T$. The residuals are defined as follows:
\begin{itemize}
    \item \textbf{Point-to-Point:}
    \begin{align}
    T^* &= \arg \min_{T \in \text{SE}(3)} \sum_{i=1}^{N} \left\| \mathbf{r}_i - T\mathbf{s}_i \right\|^2 \label{eq:pt2pt}
    \end{align}
    
    \item \textbf{Point-to-Plane:}
    \begin{align}
    T^* &= \arg \min_{T \in \mathrm{SE}(3)} \sum_{i=1}^{N} \left( \mathbf{n}_i^\top \left( \mathbf{R} \mathbf{s}_i + \mathbf{t} - \mathbf{r}_i \right) \right)^2 \label{eq:pt2pl}
    \end{align}
\end{itemize}

The high-precision odometry and updated map are integrated with Scan Context, leveraging the local LiDAR-frame cloud for loop closures, correcting accumulated drift, and improving global consistency.

\subsection{Geometric Feature Extraction}

In this section, we describe the geometric processing pipeline, which includes the spherical projection of raw point cloud data, surface normal estimation, and a per-point classification algorithm. These stages transform raw point cloud into semantically meaningful, structured representations. This enables better correspondence. This section draws inspiration from the methodology presented in \cite{normal_extraction}.
\subsubsection{Spherical Projection}

To efficiently analyse LiDAR data, we first project the raw 3D point cloud into a 2D spherical range image. Let each point be given in Cartesian coordinates as $\mathbf{p} = (x, y, z)^T$. The point is converted to spherical coordinates as follows: the range is $r = \sqrt{x^2 + y^2 + z^2}$, the azimuth angle is $\theta = \arctan(x, z)$, and the elevation angle is $\phi = \arcsin\left( \frac{y}{r} \right)$. 

We then create a 2D range image $\mathcal{I} \in \mathbb% 
{R}^{H \times W}$ by discretising the spherical field of view, where each pixel $(u, v)$ corresponds to a fixed azimuth and elevation angle.
\begin{equation} \label{eq:angular_discretization}
\begin{aligned}
\theta_u &= \theta_{\min} + u \cdot \Delta \theta, \quad \\
\phi_v   &= \phi_{\min} + v \cdot \Delta \phi
\end{aligned}
\end{equation}

where $W$ and $H$ denote the width and height of the image, and $\Delta \theta$, $\Delta \phi$ are the angular resolutions. The inverse mapping reconstructs a 3D point from the spherical pixel:
\begin{equation} \label{eq:reprojection}
\mathbf{p}_{u,v} = r_{u,v} \cdot
\begin{bmatrix}
\cos\phi_v \cdot \sin\theta_u \\
\sin\phi_v \\
\cos\phi_v \cdot \cos\theta_u
\end{bmatrix}
\end{equation}
The 3D point $\mathbf{p}_{u,v}$ corresponds to a specific spatial location in the global point cloud $\mathcal{P}$. Specifically, $\mathbf{p}_{u,v} \in \mathcal{P}$ denotes the 3D point associated with pixel coordinates $(u, v)$ in the range image. 

\subsubsection{Surface Normal Estimation}

With the spherical range image constructed, we estimate the surface normal for each valid pixel using a differential method adapted from \cite{normal_extraction}. For a pixel $(u, v)$, we define two local tangent vectors by computing the differences along horizontal and vertical directions:
\begin{equation} \label{eq:surface_tangents}
\begin{aligned}
\mathbf{t}_u &= \mathbf{p}_{u+1,v} - \mathbf{p}_{u-1,v}, \quad \\
\mathbf{t}_v &= \mathbf{p}_{u,v+1} - \mathbf{p}_{u,v-1}
\end{aligned}
\end{equation}

The surface normal is then estimated as the normalized cross product:
\begin{equation} \label{eq:normal_vector}
\mathbf{n}_{u,v} = \frac{\mathbf{t}_u \times \mathbf{t}_v}{\|\mathbf{t}_u \times \mathbf{t}_v\|}
\end{equation}
As mentioned in \cite{NV-LIOM}, to ensure a consistent orientation, the normal is flipped if it points away from the sensor:
\begin{equation} \label{eq:normal_flipping}
\text{if } \mathbf{n}_{u,v} \cdot \mathbf{p}_{u,v} > 0, \text{ then } \mathbf{n}_{u,v} \leftarrow -\mathbf{n}_{u,v}
\end{equation}

This generates a dense normal map with normal information for each pixel. This enables per-point geometric reasoning.

\subsubsection{Normal-Based Geometric Classification (Proposed)}

\begin{figure}
    \vspace{2mm}
    \centering
    \includegraphics[width=1\linewidth]{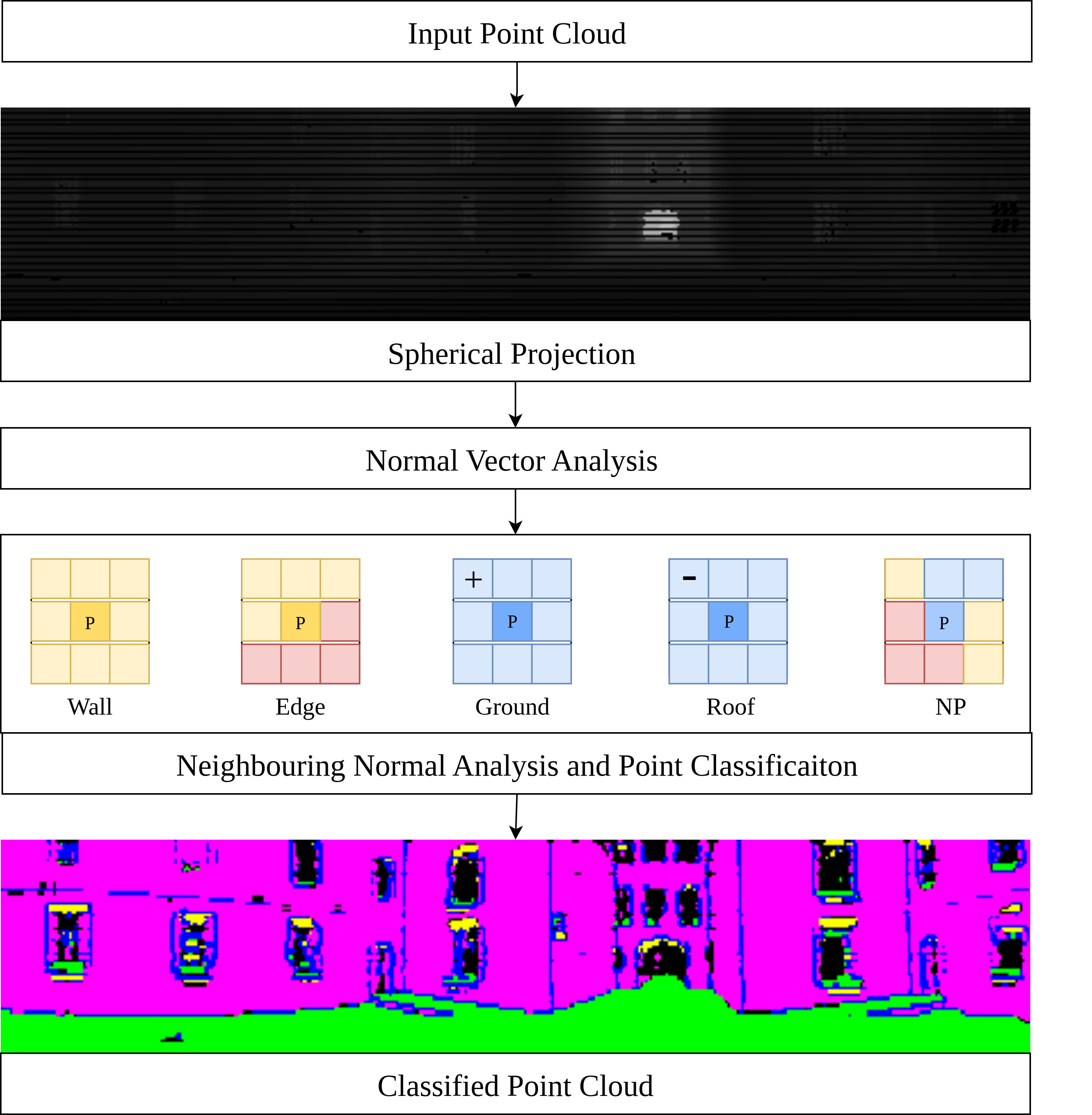}
    \caption{Overview of the geometric classification pipeline: The raw LiDAR point cloud is projected into a spherical range image, surface normals are computed using local neighborhoods, and each point is subsequently categorized into geometric classes.}
    \label{fig:image_Flow}
\end{figure}

As depicted in Fig. \ref{fig:image_Flow}, our LiDAR pipeline includes a rule-based classification algorithm. It leverages surface normals and local geometry to label each point in the LiDAR scan. 

Each point is classified into one of five categories based on its 3x3 neighbourhood (comprising the centre point and 8 neighbours) and their computed normal vectors:

\begin{itemize}
    \item Ground: Characterized by horizontal surfaces where the distribution of valid normals strongly clusters, with at least two-thirds ($\approx 66.7\%$) exhibiting a dominant Z-axis component pointing upwards ($n_z > 0$).
    \item Roof: Defined by downward-facing horizontal surfaces where at least two-thirds ($\approx 66.7\%$) of valid normals show a dominant Z-axis component pointing downwards ($n_z < 0$).
    \item Wall: Identified as vertical surfaces where at least two-thirds ($\approx 66.7\%$) of valid normals demonstrate a dominant X or Y-axis component.
    \item Edge: Detected in regions with significant normal vector divergence, specifically when the average angular variance between all pairs of valid normals in the neighbourhood exceeds a threshold of 15.0 degrees (0.26 radians).
    \item Unknown: Assigned to noisy, irregular, or ambiguous regions where valid normals are inconsistent, undefined, or fail to achieve the two-thirds majority consensus for Ground, Roof, or Wall classification.
\end{itemize}
The classified point cloud enhances semantic understanding and supports robust data association in all kinds of environments.

.
\vspace{-7mm}
\subsection{Degeneracy Aware Weighting}

In featureless or symmetrical environments, a state estimation algorithm must evaluate how well the scene geometry constrains the estimation along all translational directions. The Hessian matrix $\mathbf{H} = \mathbf{J}^\top \mathbf{J}$, where $\mathbf{J}$ is the Jacobian of the residuals, captures the curvature of the cost function with respect to the six degrees of freedom (6-DOF) of the state. The Hessian matrix $\mathbf{H}$ is partitioned into four submatrices $H'{tt}$, $H'{tr}$, $H'{rr}$, and $H'{rt}$ according to the coupling between the rotational components $\mathbf{R}$ and translational components $\mathbf{t}$ of the pose:
\begin{equation} \label{eq:hessian_block}
H' = \begin{bmatrix}
H'_{rr} & H'_{rt} \\
H'_{tr} & H'_{tt}
\end{bmatrix}_{6 \times 6}
\end{equation}

The submatrix $\mathbf{H}_{\mathbf{r}\mathbf{r}}$ and $\mathbf{H}_{\mathbf{t}\mathbf{t}}$ contain only information related to $\mathbf{R}$ and $\mathbf{t}$, and they can be used to analyse degradation in rotation and translation, respectively. 
\\
The eigenvalue analysis of $\mathbf{H}_{\mathbf{t}\mathbf{t}}$ is performed to calculate the three principal directions. The eigenvalues associated with them provide a measure of the observability in each direction.

\begin{equation} \label{eq:h_tt_eigen}
\mathbf{H}_{\mathbf{t}\mathbf{t}} = \mathbf{V}_t \boldsymbol{\Sigma}_t \mathbf{V}_t^\top 
= \mathbf{V}_t 
\begin{bmatrix}
\lambda_{t1} & 0 & 0 \\
0 & \lambda_{t2} & 0 \\
0 & 0 & \lambda_{t3}
\end{bmatrix}
\mathbf{V}_t^\top
\end{equation}

The next step involves computing the information matrix, which aggregates the contributions from all points and captures the influence of each information pair $(\mathbf{p}, \mathbf{n})$, similar to the approach proposed in~\cite{xicp}.
\begin{equation} \label{eq:information_matrix}
\mathcal{I}_t = \begin{bmatrix} \mathbf{n}_1 & \cdots & \mathbf{n}_N \end{bmatrix}^\top
\end{equation}

% All information pairs are then projected onto the eigen space of the translational Hessian by computing the dot product between the information matrix and the eigenvectors of translation. This projection yields the contribution of each point along the principal directions of translation.
All information pairs are then projected onto the eigen space of the translational Hessian. This projection yields the contribution of each point along the principal directions of translation.
\begin{equation} \label{eq:observability_score}
\mathcal{C}_t = \left| \mathcal{I}_t \cdot \mathbf{V}_t \right|
\end{equation}

The eigenvalues corresponding to each eigenvector are normalised to the range $[0, 1]$ by dividing each eigenvalue by the maximum eigenvalue. Each normalised eigenvalue is then used to scale the corresponding column of the contribution matrix $\mathcal{C}_t \in \mathbb{R}^{N \times 3}$, where each column represents the contribution of all points along a specific eigenvector direction. This column-wise weighting is given by:
\begin{equation} \label{eq:weighted_observability}
\tilde{\mathcal{C}}_t = \mathcal{C}_t \cdot 
\begin{bmatrix}
\tilde{\lambda}_{t1} & 0 & 0 \\
0 & \tilde{\lambda}_{t2} & 0 \\
0 & 0 & \tilde{\lambda}_{t3}
\end{bmatrix} \in \mathbb{R}^{N \times 3}
\end{equation}

This operation scales the contributions according to the observability of each direction. Directions with higher eigenvalues (well-constrained) receive more weight, while directions with lower eigenvalues (poorly-constrained or degenerate) receive less weight.

Subsequently, a row-wise Euclidean norm is computed to obtain a single scalar weight for each point:
\begin{equation} \label{eq:weight_computation}
\mathbf{w}_t^{(i)} = \left\| \tilde{\mathcal{C}}_t^{(i,:)} \right\|_2, \quad \text{for } i = 1, \dots, N
\end{equation}

where $\tilde{\mathcal{C}}_t^{(i,:)}$ denotes the $i^{\text{th}}$ row of the weighted contribution matrix.

This scalar weight $\mathbf{w}_t^{(i)}$ reflects how much each point contributes to localizability in the well-observable directions versus unobservable ones. 
These individual point weights can be incorporated into a weighted least-squares objective to enhance the robustness of point-cloud registration. Specifically, the residuals are defined in Section~III: System Overview, and the optimization problem is formulated as:

\begin{equation}
\min_{\mathbf{R}, \mathbf{t}} \sum_{i=1}^{N} w_i \left( \mathbf{n}_i^\top \left( \mathbf{R} \mathbf{p}_i + \mathbf{t} - \mathbf{q}_i \right) \right)^2,
\end{equation}
\vspace{2mm}
where:
\begin{itemize}
    \item \( \mathbf{p}_i, \mathbf{q}_i \) are corresponding source and target points for planar constraints,
    \item \( \mathbf{n}_i \) is the surface normal at \( \mathbf{q}_i \),
    \item \( \mathbf{R} \in \mathrm{SO}(3) \) is the rotation matrix,
    \item \( \mathbf{t} \in \mathbb{R}^3 \) is the translation vector,
    \item \( w_i \in [0, 1] \) is the degeneracy-based weight for each planar correspondence.
\end{itemize}

Apart from the degeneracy-based weighting of planar points, we also apply an adaptive Weighting scheme that accounts for the number of planar and non-planar correspondences, as introduced in~\cite{genz}. Let \( N_{\text{pl}} \) and \( N_{\text{po}} \) denote the number of planar and non-planar correspondences, respectively, and let \( N = N_{\text{pl}} + N_{\text{po}} \). The final weighted least squares objective becomes:

\begin{equation}
\hat{\mathbf{R}}, \hat{\mathbf{t}} = \arg\min 
\left[
\alpha \sum_{j=1}^{N_{\text{pl}}} \left\| e_{\text{pl}}^{(j)} \right\|^2 +
(1 - \alpha) \sum_{k=1}^{N_{\text{po}}} \left\| e_{\text{po}}^{(k)} \right\|^2
\right]
\end{equation}

where:
\begin{itemize}
    \item \( e_{\text{pl}}^{(j)} \) and \( e_{\text{po}}^{(k)} \) are the planar and non-planar residuals (defined in Section~III),
    \item \( \alpha \in [0, 1] \) is the adaptive weight based on the relative number of planar and non-planar correspondences.
\end{itemize}
\vspace{-0 cm}

\subsection{Scan-Context Back-end}
To improve global consistency and enable loop closure, we integrate the Scan Context \cite{scan_context} algorithm into the back-end of our SLAM system. For each incoming 3D LiDAR scan, a grid-based polar descriptor is generated by encoding the maximum height of points within discretised azimuthal and radial bins. These descriptors are stored and indexed using a KD tree based on a rotation-invariant ring key. During run-time, the descriptor of the current scan is matched against previous entries to identify potential loop closures using a two-phase retrieval process involving coarse nearest-neighbour search followed by fine-grained similarity scoring using column-wise cosine distance. Upon successful loop detection, the corresponding pose pair is added as a loop constraint in the pose graph, enabling back-end optimization to correct accumulated drift and refine the global map.
\section{Results}

To evaluate our method, we used the widely used EVO evaluator~\cite{evo}, which provides metrics for both absolute and relative pose errors. We tested our algorithm under consistent parameters across a variety of challenging datasets, including degenerate environments and outdoor sequences. Our approach demonstrated significantly improved performance compared to current state-of-the-art algorithms, particularly in the absolute pose error metric. We consistently outperformed existing methods on all datasets in this metric. Although in the relative pose error metric, some baseline methods showed competitive or superior results in specific sequences, our algorithm still achieved highly competitive overall performance.
\vspace{4mm}  % adjust as needed: 3mm or 5mm

\noindent
\begin{table}[htbp]
	\centering
	\setlength{\tabcolsep}{2.3pt}
    \captionsetup{font=footnotesize}
	\caption{Quantitative results for the Long\_Corridor sequence of the SubT-MRS dataset~\cite{Subt}. Some comparison results have been reproduced from ~\cite{genz}.}

{\scriptsize
\setlength{\tabcolsep}{1.5pt} % <-- Reduced column spacing
\begin{tabular}{c|cccc|cccc}
    \toprule \midrule
    \multirow{2}{*}{\textbf{Method}} & \multicolumn{4}{c|}{\textbf{Absolute pose error [m]}} & \multicolumn{4}{c}{\textbf{Relative pose error [m]}} \\ 
    \cmidrule(lr){2-5} \cmidrule(lr){6-9}
    & \textbf{Mean} & \textbf{Max} & \textbf{RMSE} & \textbf{Stdev.} & \textbf{Mean} & \textbf{Max} & \textbf{RMSE} & \textbf{Stdev.} \\ 
    \midrule  
    KISS-ICP~\cite{KISS-ICP} & 6.83 & 19.05 & 8.72 & 5.41 & 0.10 & 0.94 & 0.14 & 0.10 \\
    CT-ICP~\cite{cticp} & 44.18 & 60.14 & 45.66 & 11.55 & 0.19 & 7.15 & 0.68 & 0.65 \\
    DLO~\cite{DLO} & 7.69 & 27.99 & 9.09 & 4.86 & 0.26 & 22.74 & 1.32 & 1.29 \\ 
    Point-to-point ICP~\cite{pointtopoint} & 6.83 & 19.05 & 8.72 & 5.41 & 0.10 & 0.94 & 0.14 & 0.10 \\
    Point-to-plane ICP~\cite{planetoplane} & 32.84 & 40.88 & 33.16 & 4.55 & 0.13 & 12.50 & 0.68 & 0.67 \\ 
    GENZ-ICP~\cite{genz} & 1.69 & 4.32 & 1.99 & 1.04 & \textbf{0.06} & \textbf{0.73} & \textbf{0.09} & \textbf{0.07} \\ 
    \midrule
    \textbf{Ours} & \textbf{1.47} & \textbf{4.08} & \textbf{1.72} & \textbf{0.89} & 0.12 & 0.97 & 0.17 & 0.13 \\
    \midrule \bottomrule
\end{tabular}
}

\end{table}
  % Forces LaTeX to stop here and not float the next table above

 % Prevent the second table from jumping above the first% Prevent the second table from jumping above the first
\begin{table}[H]
	\centering
	\setlength{\tabcolsep}{1.7pt}
    \captionsetup{font=footnotesize}
	\caption{Quantitative results for the Corridor1 and Corridor2 sequences of the Ground-Challenge dataset~\cite{ground_challenge}. Some comparison results have been reproduced from~\cite{genz}}

    {\scriptsize
	\begin{tabular}{c|c|cccc|cccc}
		\toprule \midrule
		\multirow[c]{3}{*}{\textbf{Sequence}} & \multirow{3}{*}{\textbf{Method}} & \multicolumn{4}{c|}{\textbf{Absolute pose error [m]}} & \multicolumn{4}{c}{\textbf{Relative pose error [m]}} \\ \cmidrule(lr){3-6} \cmidrule(lr){7-10}
		& & \textbf{Mean} & \textbf{Max} & \textbf{RMSE} & \textbf{Stdev.} & \textbf{Mean} & \textbf{Max} & \textbf{RMSE} & \textbf{Stdev.} \\ \midrule
		\multirow[c]{7.5}{*}{\shortstack{Corridor1\\(zigzag)}}     
		& KISS-ICP~\cite{KISS-ICP} & 1.70 & 4.76 & 2.17 & 1.35 & 0.12 & 0.59 & 0.15 & 0.09 \\
		& CT-ICP~\cite{cticp} & 0.44 & 1.05 & 0.54 & 0.30 & 0.05 & \textbf{0.23} & \textbf{0.06} & 0.04 \\
		& DLO~\cite{DLO} & 0.34 & 1.04 & 0.45 & 0.30 & 0.05 & 0.55 & 0.08 & 0.06 \\
		\cmidrule(lr){2-10}
		& Zhang~\cite{zhang} & 0.22 & 0.67 & 0.28 & 0.17 & 0.05 & 0.26 & 0.06 & 0.04 \\
		& X-ICP~\cite{xicp} & 0.94 & 9.45 & 2.05 & 1.83 & 0.05 & 0.47 & 0.07 & 0.05 \\
		& GENZ-ICP~\cite{genz} & 0.19 & 0.49 & 0.24 & 0.14 & \textbf{0.04} & \textbf{0.23} & 0.06 & 0.04 \\ \midrule
        & Ours & \textbf{0.05} & \textbf{0.17} & \textbf{0.06 } & \textbf{0.02} & \textbf{0.04} & 0.46 & \textbf{0.05} & \textbf{0.03} \\ \midrule
		\multirow[c]{8}{*}{\shortstack{Corridor2\\(straight\\ forward)}} 
		& KISS-ICP~\cite{KISS-ICP} & 0.54 & 1.34 & 0.68 & 0.41 & 0.14 & 0.46 & 0.16 & 0.08 \\
		& CT-ICP~\cite{cticp} & 1.04 & 2.36 & 1.30 & 0.78 & \textbf{0.12} & 0.35 & \textbf{0.14} & \textbf{0.06} \\
		& DLO~\cite{DLO} & 0.72 & 1.72 & 0.93 & 0.59 & \textbf{0.12} & \textbf{0.34} & \textbf{0.14} & 0.07 \\ 
		\cmidrule(lr){2-10}
		& Zhang~\cite{zhang} & 0.21 & 0.60 & 0.28 & 0.18 & \textbf{0.12} & 0.36 & \textbf{0.14} & \textbf{0.06} \\
		& X-ICP~\cite{xicp} & 5.85 & 9.69 & 6.92 & 3.70 & 0.13 & 0.37 & 0.15 & 0.07 \\
		& GENZ-ICP~\cite{genz} & 0.18 & 0.41 & 0.20 & 0.09 & \textbf{0.12} & 0.36 & \textbf{0.14} & 0.07 \\
		\midrule
        & Ours & \textbf{0.07} & \textbf{0.21} & \textbf{0.08} & \textbf{0.04} & \textbf{0.12} & 0.39 & \textbf{0.14} & \textbf{0.06} \\
		\midrule \bottomrule
	\end{tabular}
    }
    \label{table:ground}
\end{table}
\begin{table}[H]
	\centering
	\setlength{\tabcolsep}{2.4pt}
    \captionsetup{font=footnotesize}
	\caption{Quantitative results for the  Newer College dataset~\cite{newer_college}.}
	{\scriptsize
	\begin{tabular}{c|cccc|cccc}
		\toprule \midrule
		\multirow{2}{*}{\textbf{Method}} & \multicolumn{4}{c|}{\textbf{Absolute pose error [m]}} & \multicolumn{4}{c}{\textbf{Relative pose error [m]}} \\ \cmidrule(lr){2-5} \cmidrule(lr){6-9}
		& \textbf{Mean} & \textbf{Max} & \textbf{RMSE} & \textbf{Stdev.} & \textbf{Mean} & \textbf{Max} & \textbf{RMSE} & \textbf{Stdev.} \\ \midrule  
		KISS-ICP~\cite{KISS-ICP} & 0.16 & 0.49 & 0.24 & 0.09 & 0.22 & 0.78 & 0.24 & 0.09 \\
		 GENZ-ICP~\cite{genz} & 0.14 & 0.43 & \textbf{0.15} & 0.07 & \textbf{0.22} & \textbf{0.78} & \textbf{0.24} & \textbf{0.09} \\ \midrule
        \textbf{Ours} & \textbf{0.13} & \textbf{0.35} & \textbf{0.15} & \textbf{0.06} & 0.49 & 2.26 & 0.68 & 0.48 \\
		\midrule \bottomrule
	\end{tabular}
    }
    \label{table:iccv}
    \vspace{-0.0cm}
\end{table}

 % Forces LaTeX to stop here and not float next table above
% Table 2: Ground-Challenge Dataset Results
% \subsection*{Limitations}

% While the proposed method offers fast and efficient geometric classification, it has certain limitations:

% \begin{enumerate}
%     \item \text{Sensor Dependence:} The current pipeline is designed for rotating (repeating) LiDARs with structured scan patterns. It does not generalise well to pattern-based or solid-state LiDARs, where the range image formation is irregular or non-uniform.
    
%     \item \text{Sensitivity in Unstructured Environments:} Although the segmentation algorithm is computationally efficient, it may produce outliers in highly unstructured or irregular environments such as caves, where normal estimation becomes unreliable due to sparse or noisy local geometry.
% \end{enumerate}

\section{CONCLUSIONS}
In this work we proposed DAMM-LOAM, a degeneracy-aware LiDAR odometry and mapping system designed to improve pose estimation in challenging, low-feature environments. Our approach classifies points based on geometric features derived from the normal map and leverages this classification to assign semantic-specific weights to correspondences. A novel observability-aware weighting scheme, computed using the translational Hessian eigenvalues, further enhances robustness by reducing the influence of poorly observable directions during optimization. Experimental results on multiple degenerate and outdoor datasets demonstrate that DAMM-LOAM shows significant improvements in absolute pose accuracy.
\newline
\newline
However, since our approach is dependent on the range and normal map, it is currently limited to repeating LiDARs only. Although the segmentation algorithm is computationally efficient, it may produce outliers in highly unstructured or irregular environments.

Future work will explore fusion with visual sensors to improve feature segmentation and extend degeneracy analysis to rotational components. Learning-based classification from normal maps is another promising direction to improve generalization across diverse environments.

\addtolength{\textheight}{-0cm}   % This command serves to balance the column lengths

%%%%%%%%%%%%%%%%%%%%%%%%%%%%%%%%%%%%%%%%%%%%%%%%%%%%%%%%%%%%%%%%%%%%%%%%%%%%%%%%

%%%%%%%%%%%%%%%%%%%%%%%%%%%%%%%%%%%%%%%%%%%%%%%%%%%%%%%%%%%%%%%%%%%%%%%%%%%%%%%%

%%%%%%%%%%%%%%%%%%%%%%%%%%%%%%%%%%%%%%%%%%%%%%%%%%%%%%%%%%%%%%%%%%%%%%%%%%%%%%%%

\bibliographystyle{IEEEtran}  % Choose appropriate style
\bibliography{root}     % .bib file containing the 

\end{document}